# Design of Mobile Manipulator for Fire Extinguisher Testing. Part I: Key Specifications and Conceptual Design

Xuan Quang Ngo[1], Thai Nguyen Chau[1], Cong Thang Doan[1], Van Tu Duong[123], Duy Vo Hoang[4] and and Tan Tien Nguyen[123(✉)]

[1]National Key Laboratory of Digital Control and System Engineering (DCSELab), Ho Chi Minh City University of Technology (HCMUT), 268 Ly Thuong Kiet Street, District 10, Ho Chi Minh City, Viet Nam
[2]Faculty of Mechanical Engineering, Ho Chi Minh University of Technology (HCMUT), 268 Ly Thuong Kiet, District 10, Ho Chi Minh City, Viet Nam
[3]Vietnam National University Ho Chi Minh City, Linh Trung Ward, Thu Duc District, Ho Chi Minh City, Viet Nam
[4]Faculty of Electrical and Electronics Engineering, Ton Duc Thang University, Ho Chi Minh City, Viet Nam
nttien@hcmut.edu.vn

**Abstract.** All flames are extinguished as early as possible, or fire services have to deal with major conflagrations. This leads to the fact that the quality of fire extinguishers has become a very sensitive and important issue in firefighting. Inspired by the development of automatic fire fighting systems, this paper proposes key specifications based on the standard of fire extinguishers that is ISO 7165:2009 and ISO 11601:2008, and feasible solutions to design a mobile manipulator for automatically evaluating the quality or, more specifically, power of fire extinguishers. In addition, a part of the mechanical design is also discussed.

## 1 Introduction

Nowadays, there has been increasing in the study on automatic fire fighting systems because firefighting is a risky profession. For instance, Chee Fai Tan et al. described different types of current fire fighting machines, which are composed of a rubber track system, a nozzle system blowing an extremely high-pressure water beam, thermal imaging cameras, and sensors for observation and monitoring purpose in fire scene [1]. Poonam Sonsale et al. studied the observation system in intelligent buildings, which detects fire, alarm people, and automatically transfers fire extinguishers to the hot spot [2]. The nozzle, alarm and detection system and fire extinguishers are used for the purpose of putting out flames. As a result, the equipment is an indispensable component of the fire fighting system. Therefore, a fire extinguisher of any random batch is required to evaluate and meet quality standards for the fire extinguishers mentioned in ISO 7165:2009 for Fire fighting – Portable fire extinguishers – Peformance and construction and ISO 11601:2008 Fire fighting – Wheeled fire extinguishers – Performance and contruction.

The standards propose different classes of fire tests, such as class A, class B, class C and class K fire test whose characteristics and dimensions are a means of determining power of fire extinguishers. For instance, MFZL8-ABC with the power of 4A as mentioned in Table 1, means that it is able to extinguish the class A fire test having a bigger dimension than the one MFZL4-ABC with the power of 2A can do. So far, it cannot find studies on fire extinguisher testing automation. This task still has been mainly performed by humans. Consequently, firefighters could be faced with potentially life-threating situations such as explosions and firefighters' poor health resulting from working in a hard environment for long hours, such as high temperatures, dust, and humidity. In order to contribute to the development of automating the fire fighting system, this paper studies a mobile manipulator for testing fire extinguishers based on ISO 7165:2009 and ISO 11601:2008.

There are some common fire extinguishers in a market with types of corresponding fire tests, as shown in Table 1, and the process for using the fire extinguisher to put down the fire includes four steps as shown in Fig. 1.

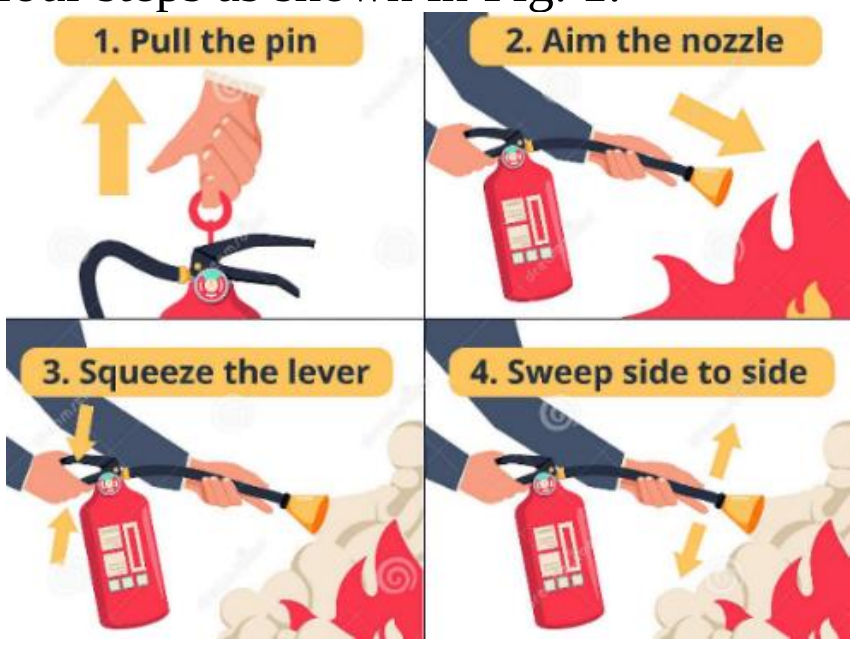

**Fig. 1.** The process of performing the fire extinguisher

**Table 1.** Some common fire extinguishers with types of corresponding fire test

| **Model** | **EXT-ABC-4K** | **MFZL4-ABC** | **MFZL8-ABC** | **MFZL10-ABC** |
|---|---|---|---|---|
| Type | Portable | Portable | Portable | Wheeled |
| Power | 21A/ 133B | 2A/ 55B | 4A/ 89B | 20A/ 233B |
| Weight | $6.1\ kg$ | $5.5\ kg$ | $10\ kg$ | $45\ kg$ |
| Dimension | $\phi 13.8 \times 44\ cm$ | $\phi 13 \times 48\ cm$ | $\phi 13 \times 56.5\ cm$ | $\phi 46 \times 92\ cm$ |
| Length of hose | $50\ cm$ | $40\ cm$ | $50\ cm$ | $3\ m$ |
| Discharge time | $15\ s$ | $13\ s$ | $15\ s$ | $20\ s$ |
| Operating temp. | $-20^oC - 60^oC$ | $-20^oC - 55^oC$ | $-20^oC - 55^oC$ | $-20^oC - 55^oC$ |
| **Model** | **EXT-CO2-5K** | **CO2-MT24** | **EXT-ABC-25K** | **EXT-ABC-50K** |
| Type | Portable | Wheeled | Wheeled | Wheeled |
| Power | 55B | 233B | 20A/ 89B | 20A/ 233B |
| Weight | 16.8 $kg$ | 90 $kg$ | 50 $kg$ | 83 $kg$ |
| Dimension | $\phi 15.2 \times 67\ cm$ | $\phi 22 \times 133\ cm$ | $\phi 25.2 \times 88\ cm$ | $\phi 30 \times 100\ cm$ |
| Length of hose | 50 $cm$ | 3 $m$ | 5 $m$ | 5 $m$ |
| Discharge time | 15 $s$ | 25 $s$ | 20 $s$ | 25 $s$ |
| Operating temp. | $-20^oC - 60^oC$ | $-20^oC - 55^oC$ | $-20^oC - 60^oC$ | $-20^oC - 60^oC$ |

According to the implementation process of testing the fire extinguishers from ISO 7165:2009 and ISO 11601:2008, based on the operation of the fire extinguisher and the specification of commercial fire extinguishers, this paper proposes key specifications in Table 2. Note that, since the dimension of the class C, D, and K fire tests are covered in the class A fire test, only the fire tests class A and B are mentioned in this study, and the discharge nozzle of the fire extinguisher is assumed to be held by the end-effector of a mobile manipulator.

**Table 2.** Key specifications to design the mobile manipulator for fire extinguisher testing

| **Requirements of load** | | |
|---|---|---|
| | Portable fire extinguishers | Wheeled fire extinguishers |
| Mass | 20 $kg$ | 100 $kg$ |
| Length of hose | 40 $cm$ to 60 $cm$ | 3 $m$ to 5 $m$ |
| Dimension | $\phi 185\ mm \times 700\ mm$ | $\phi 350\ mm \times 1300\ mm$ |
| Length of nozzle | 50 $mm$ to 500 $mm$ | … |
| **Requirements of operating range** | | |
| Class A fire test, power 20A, $1270 \times 1270 \times 1725\ mm$ (length × width × height) as shown in Fig. 2A | | |
| Class B fire test, power 233B, $\phi 3000 \times 203\ mm$ (diameter × height) as shown in Fig. 2B | | |
| | Class A fire test | Class B fire test |
| The highest position of the end-effector | 1892 $mm$ (point A) | … |
| The lowest position of the end-effector | 745 $mm$ (point B) | 635 $mm$ (point C) |
| The shortest distance between the centroid of the robot and the outer edge of the fire test | 1698 $mm$ | 1338 $mm$ |
| The shortest distance between the end-effector and the outer edge of the class A fire test | 395 $mm$ | … |
| The longest distance between the end-effector and the centroid of the robot | 1875 $mm$ (point A) | 2255 $mm$ (point D) |
| **Other requirements** | | |

| | |
|---|---|
| Operating temperature | $-10^oC$ to $55^oC$ |
| Time | Total time for moving aroung the fire test must be lesser the discharge time of the fire extinguisher |
| Mobility | The robot has the ability to climb a 30-degree slope and stairs, the comparable power when moving forward and backward and move through soft soil |

In section 2, several solutions are proposed to deal with the above problems. Section 3 presents the process of designing the mobile manipulator based on the solution. In section 4, works presented in part II of this study are discussed.

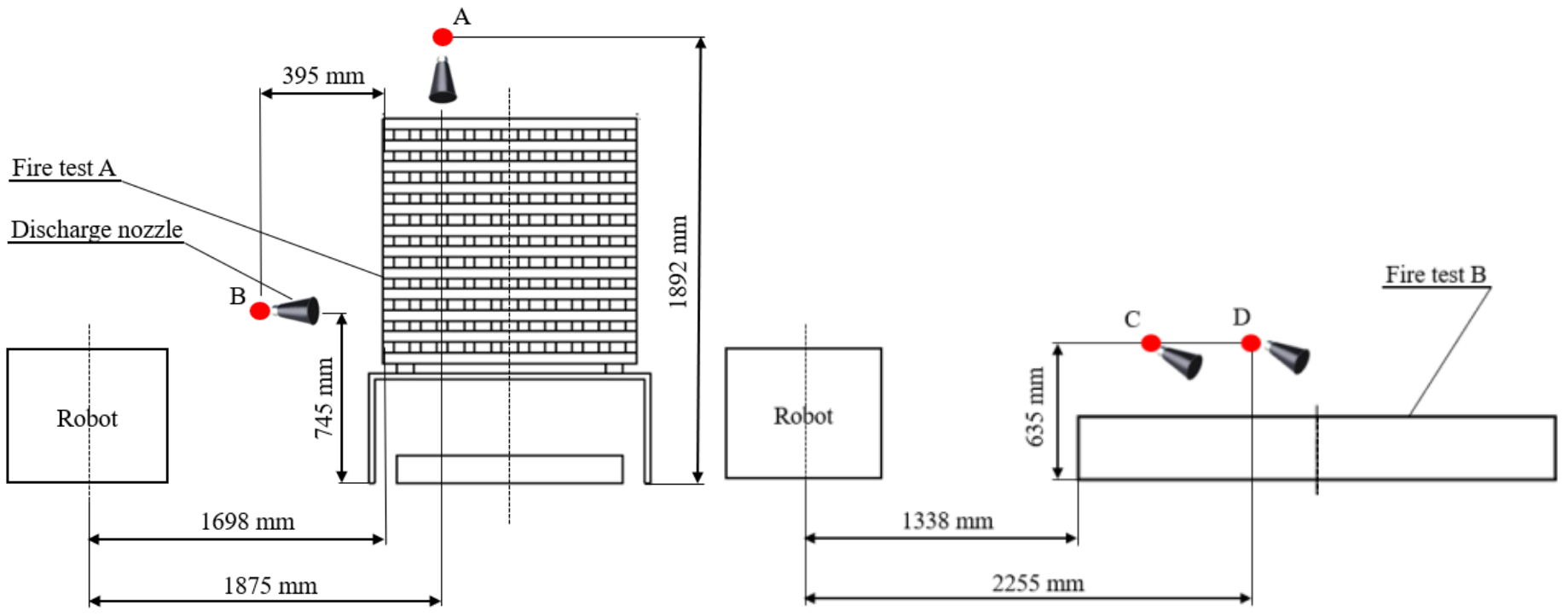

a) For the class A fire test b) For the class B fire test

**Fig. 2.** Requirements of the robot's operating range

## 2 Conceptual Design

In this section, the paper proposes two conceptual designs to simulate human activity by using fire extinguishers to extinguish the fire test.

## 2.1 Rail-Track Manipulator

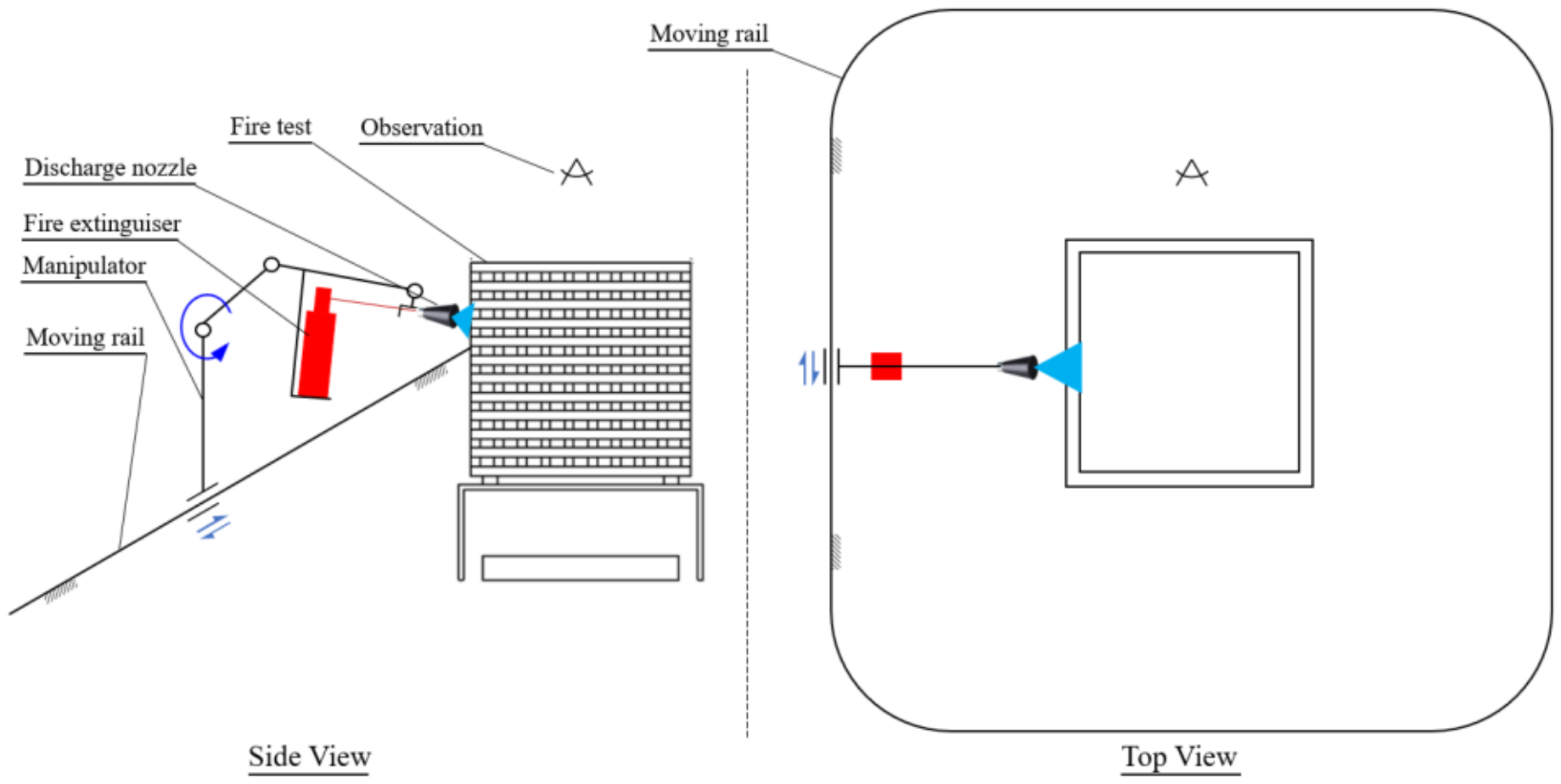

**Fig. 3.** Rail-track manipulator to test the fire extinguishers automatically

As shown in Fig. 3, a manipulator carrying the fire extinguisher moves around the fire test on a moving rail and aims the discharge nozzle in the direction of the target through the processing signal of the fire detecting camera or sensor installed at the end-effector of the manipulator. This solution is able to simplify the control of the mobile manipulator's movement around the fire test. However, it places a great burden on the manipulator's reach due to the fixed distance between the moving rail and the fire test, and there are extra constructions in the testing area.

## 2.2 Mobile Manipulator

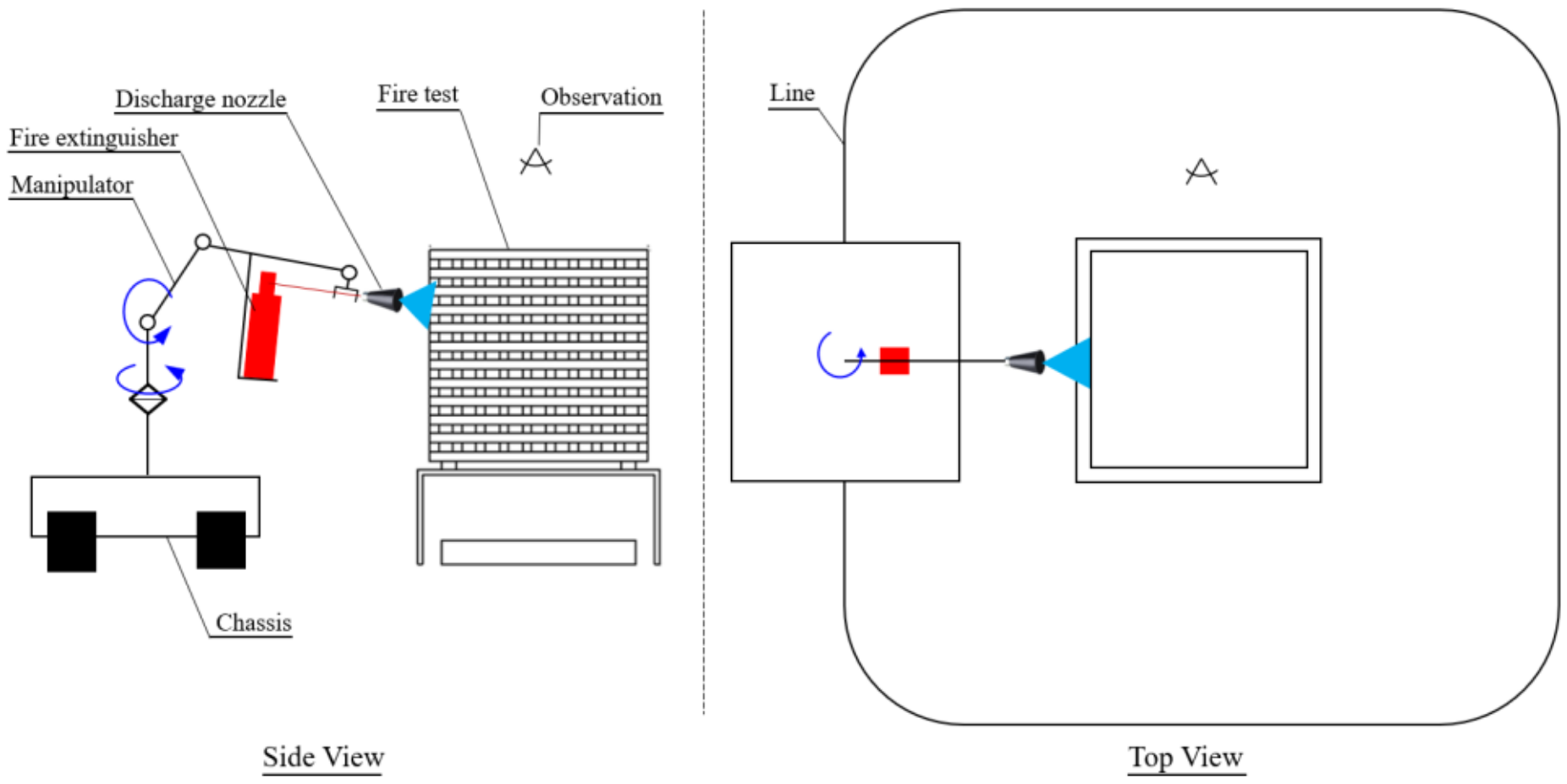

**Fig. 4.** Mobile manipulator to test the portable fire extinguishers automatically

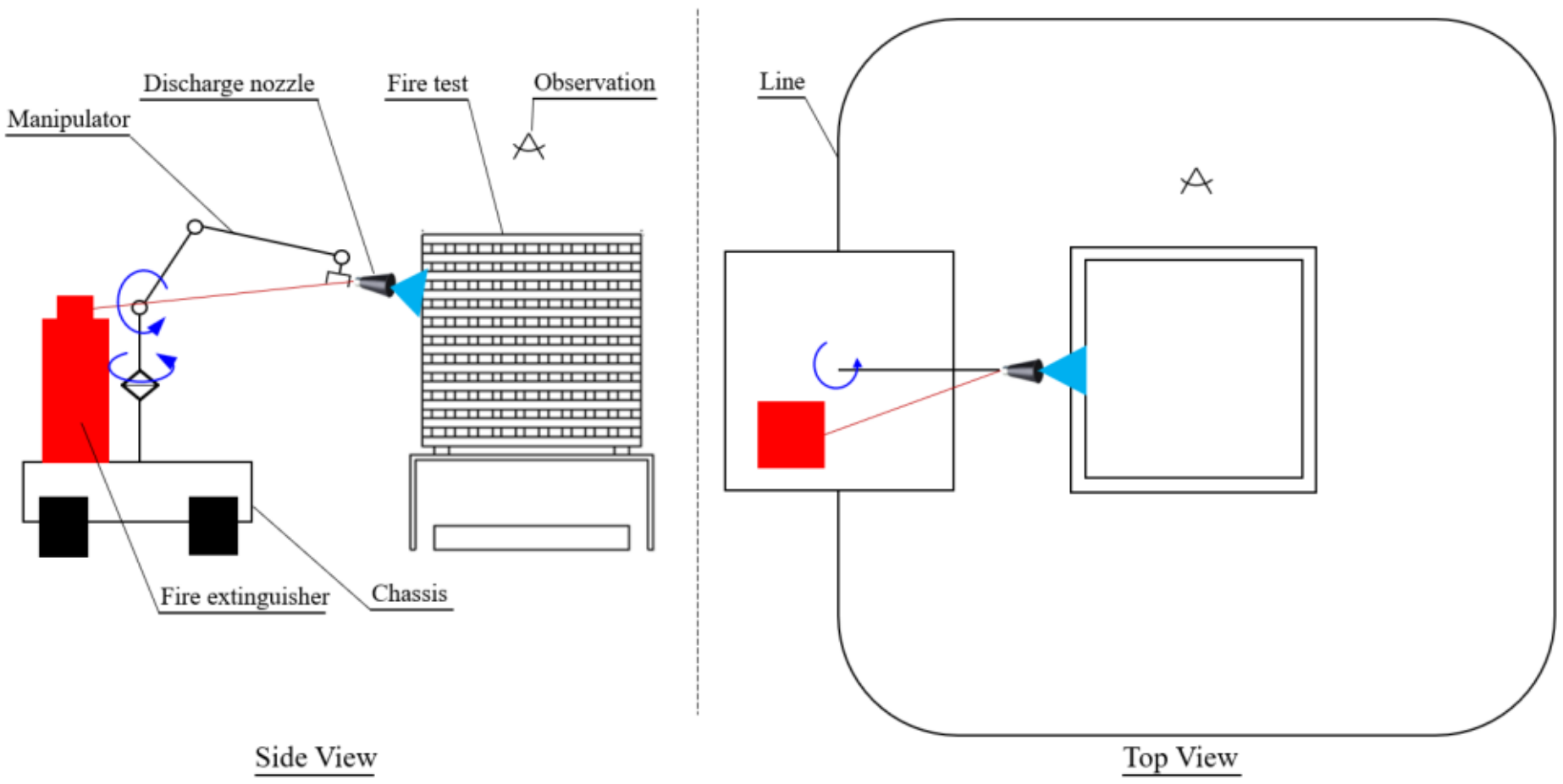

**Fig. 5.** Mobile manipulator to test the wheeled fire extinguishers automatically

A manipulator is placed on a moving chassis tracking a known trajectory around the fire test, as shown in Fig. 4 and Fig. 5. Although this system requires designing and controlling the chassis, there are no additional constructions in the testing area and the responsibility for the manipulator's reach could be mitigated by modifying the line. Thus, the mobile manipulator is selected.

**Types of Chassis.** According to J.Y.Wong and Wei Huang [3], who studied the performance of wheeled chassis and tracked chassis on various soil, the thrust developed by the wheeled chassis with lower shear stress and the lower contact area is generally lower than that developed by the comparable tracked chassis. In addition, the ability to climb stairs is also considered. This leads to the fact that the tracked chassis is well-preferred instead of the wheeled chassis in terms of rough terrain movement.

**Material of The Track.** Although steel tracks have some advantages, features of rubber tracks such as good traction on smooth and dry surfaces, low vibrations, low noises, and no damage to the concrete and asphalt surface are suitable of the application of testing the fire extinguisher. Thus, the rubber track is suitable for the application.

**Shape of The Track.** Thanks to the advance properties of the tracks which are similar to a trapezoid in shape as shown in Fig. 8, so they are preferable in terms of the comparable capability of moving forward and backward than the tracks having a rhomboid shape with huge advantages of crossing large tranches.

**Manipulator Configuration.** The task of testing the fire extinguisher requires the robot to possess 5 degrees of freedom (DOFs) with the kinematic diagram as presented in Fig. 6. Joint A of the manipulator is a revolute joint rotating around the $z$-axis. Joints B and C are also revolute joints so that the manipulator can direct the end-effector and the discharge nozzle to the fire like the step 2, as shown in Fig. 1. A revolute joint is added to joint G, which helps to control the orientation of the discharge nozzle more conveniently. In addition, in order to simulate step 4, as shown in Fig. 1, the slider-crane mechanism is added to sweep the nozzle discharge side to side, as shown in Fig.

7. The parallelograms forming the closed kinematic loops allow to put driving motors on the base and enhancing the total stiffness, and thus it reduces the inertia.

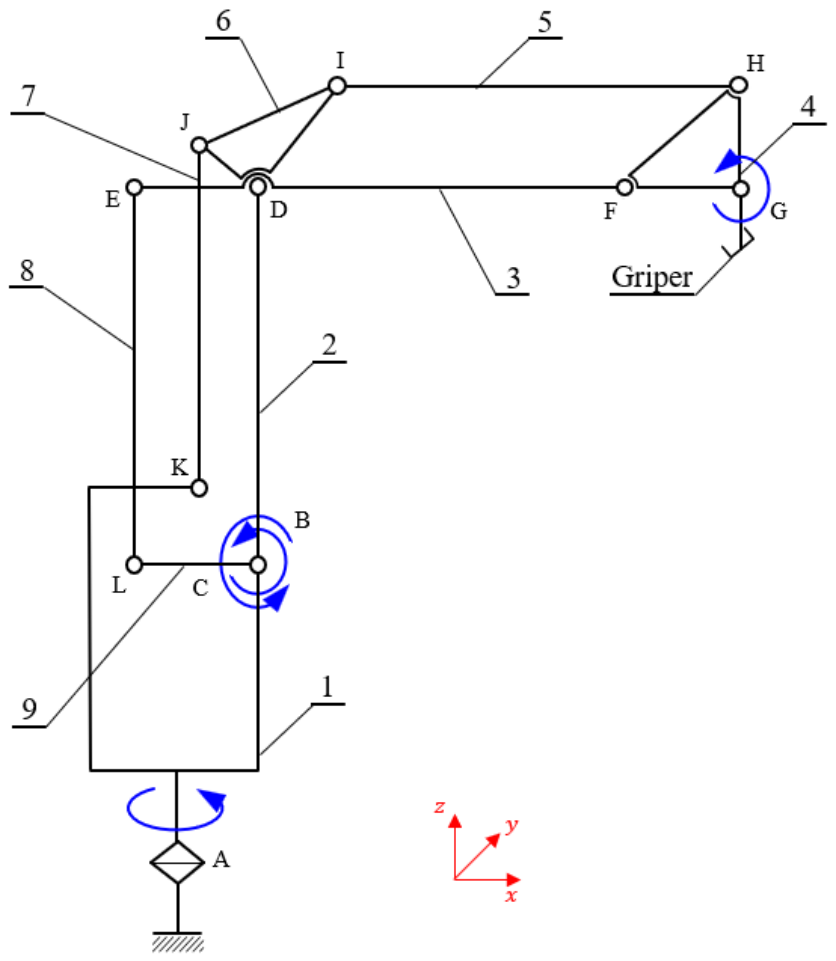

**Fig. 6.** The kinematic diagram of the palletizing manipulator – Side view

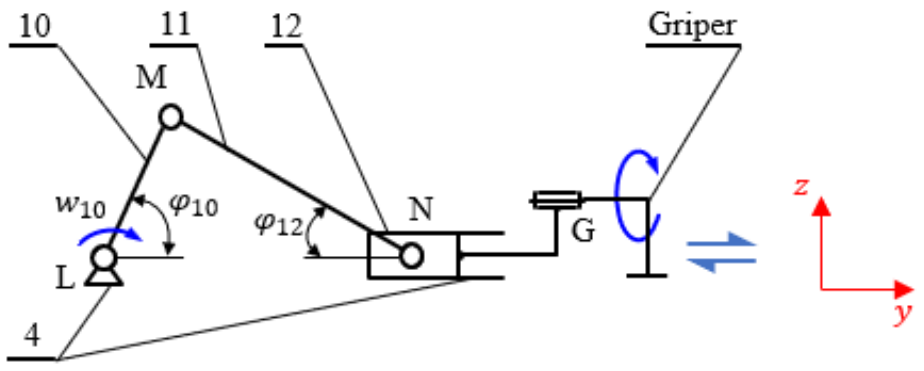

**Fig. 7.** The slider-crane mechanism sweeps the gripper side to side – Front view

**Types of manipulator's motors.** Servo motors are applied widely for the palletizing manipulator [9], due to their compact, incredible energy efficiency, and repeatability of motion. Thus, servo motors are used for the manipulator design.

## 3 Chassis Design

The speed fluctuation $\delta$ due to the polygon effect of the sprocket tooth-track link engagement, which should fall into the range of 72 – 2.75%, is given by [4]:

$$\delta = 1 - \sqrt{1 - \left(\frac{P}{D}\right)^2} \quad (1)$$

where, $P$ is the track pitch; $D$ is the pitch diameter of the sprocket.

The ratio of roadwheel spacing and track pitch $r_{RSP}$ should be about 1.5 to 1.8 to achive the best tractive efficiency for low speed tracked vehicles [4]:0 above

$$r_{RSP} = \frac{RS}{P} \quad (2)$$

where, $RS$ is the roadwheel spacing.

The ratio of roadwheel diameter and track pitch $r_{RDP}$ should be about 1.2 for low-speed tracked vehicles [5]:

$$r_{RDP} = \frac{RD}{P} \tag{3}$$

where, $RD$ is the roadwheel diameter.

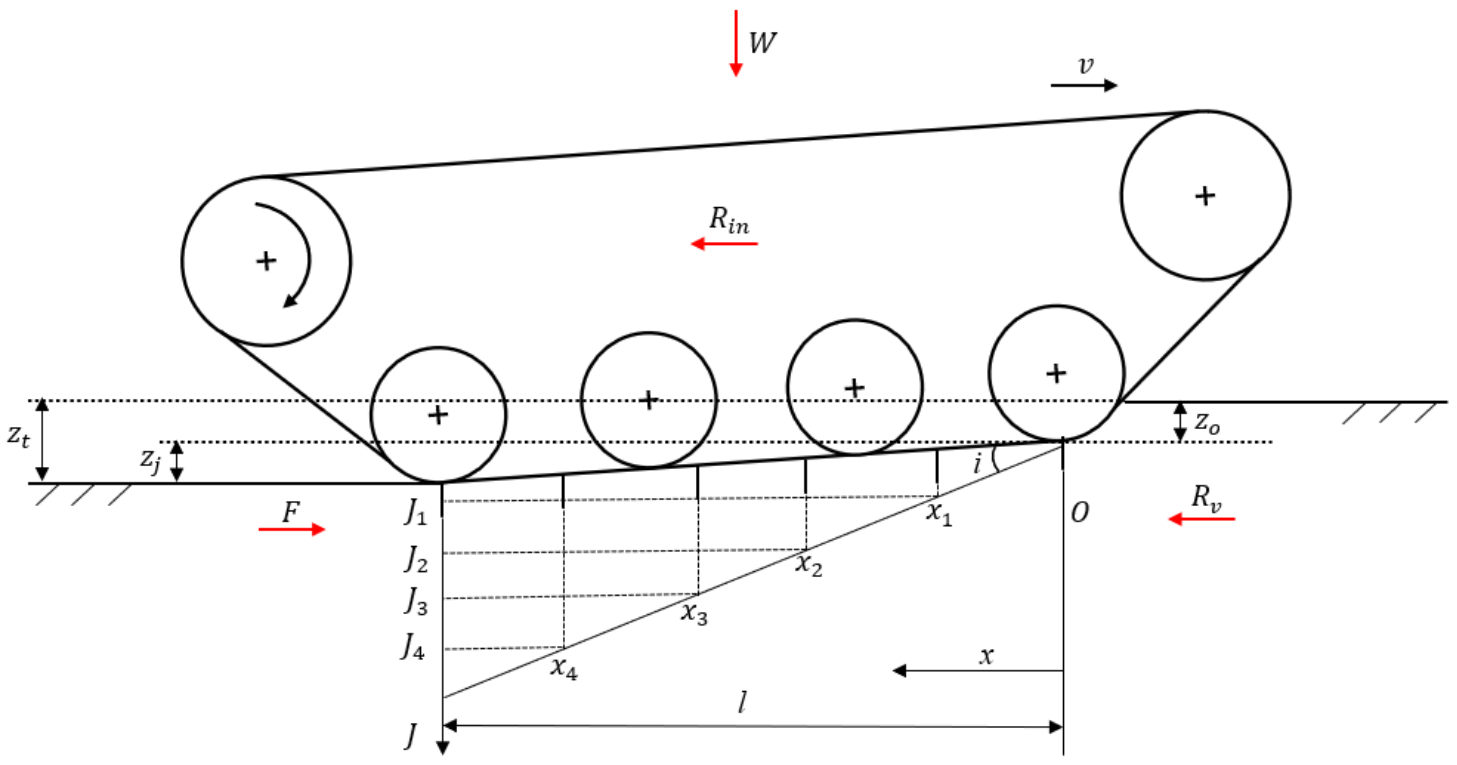

**Fig. 8.** Major external forces acting on a track chassis

According to Newton second law, the motion of the tracked chassis is illustrated by:

$$ma = F - R_b - R_c - R_{in} - R_g \tag{4}$$

where, $F$ is the sum of thrusts; $R_b$ is the bulldozing resistance caused by the static sinkage $z_o$; $R_c$ is the compaction resistance caused by the static sinakge $z_o$ and the slip sinkage $z_j$; $R_{in}$ is the internal resistance of the running gear; $R_g$ is the grade resistance; $m$ is the robot mass; $a$ is the acceleration.

The average value of the internal resistance $R_{in}$ of tracked vehicles coming from the frictional losses in track pins, between the driving sprocket teeth and track links, the rolling resistance of the roadwheels on the track, etc. is estimated by [5]:

$$R_{in} = \frac{W}{1000}(133 + 9v) \tag{5}$$

where, $W$ is vehicle weight; $v$ is the vehicle speed.

The bullzoding resistance $R_b$ is calculated by [6]:

$$R_b = 2b\int_0^{z_0}\left(\gamma K_p z + 2c\sqrt{K_p}\right)dz \tag{6}$$

where, $b$ is the track system width; $K_p$ is the Rankine passive earth pressure coefficient determined by the internal friction angle $\phi$ of the soil, $K_p = \tan^2(\pi/4 + \phi/2)$ [8]; $c$ is the cohesion of the terrain; $\gamma$ is the unit weight of the soil.

The static sinkage $z_o$ is estimated based on the ground contact pressure $P$ [7]:

$$z_o = \left(\frac{p}{k_c/b + k_\phi}\right)^{1/n} \tag{7}$$

where, $p$ is the normal pressure assumed to be uniform distribution, $p = W/2bl$.

The compaction resistance $R_c$ is calculated by [6]:

$$R_c = \frac{bkz_o^{n+1}}{l(n+1)} \int_0^l \left(78 - 2.78e^{-0.009(ix)^{1.77}}\right)^{n+1} dx \tag{8}$$

where, $l$ is the soil-track contact length; $k$ is the sinkage modulus, $k = k_c/b + k_\phi$ [7]; $n, k_c$ and $k_\phi$ are parameters of terrain; $x$ is distance from the front of the contact area.

An off-road tracked vehicle's tractive performance does not only rely on its engine power, but also is limited by the soil thrust determined from the soil-track interaction, which is calculated by [5]:

$$F = (Ac + W \tan\phi)\left[1 - \frac{K}{il}\left(1 - e^{-il/K}\right)\right] \tag{9}$$

where, $F$ is the total tractive effort of a track; $A$ is the ground contact area; $K$ is the shear deformation modulus; $i$ is slip ratio.

The ratio of track length to treat of the chassis, $l/B$, must satisfy Eq. (10) to ensure the tracked chassis to steer without spinning the outside track [5]:

$$\frac{l}{B} \leq \frac{2}{\mu_t}\left(\frac{c}{p} + \tan\phi - f_r\right) \tag{10}$$

where, $f_r$ is the coefficient of motion resistance of the vehicle in the longitudinal direction; $\mu_t$ is the coefficient of lateral resistance.

The total mass, $m$, put on the chassis is approximately $300\ kg$ including the mass of the fire extinguisher, the manipulator, and other components. Using Eq. (1) to (10) and solving for the tractive effort and the acceleration of the chassis as shown in **Table 3**, the results show that the performance of the tracked chassis with assumed parameters as shown in Table 3 is able to fulfil the requirements of the key specifications.

**Table 3.** Tractive effort and acceleration of the chassis

| | | | | |
|---|---|---|---|---|
| Soft soil parameters [4] | $n = 0.8$ | $k_c = 16.54\ kN/m^{1.5}$ | $k_\phi = 911.4\ kN/m^{2.5}$ | |
| | $\phi = 29^o$ | $\gamma = 15\ kN/m^3$ | $K = 2.5\ cm$ | |
| | $c = 6.89\ kPa$ | $i = 0.2$ [6] | | |
| Assumed parameters of chassis | $B = 0.8\ m$ | $m = 300\ kg$ | $v = 1.5\ m/s$ | $D = 0.18\ m$ |
| Assmued parameters of track | $b = 0.18\ m$ | $l = 0.1\ m$ | $P = 0.155\ m$ | |
| Speed fluctuation | $\delta = 62\ \%$ | | | |
| Roadwheel parameters | $RS = 0.23\ m$ | $RD = 0.19\ m$ | | |
| Internal resistance | $R_{in} = 431.2\ N$ | | | |
| Motion resistance | $z_o = 0.0024\ m$ | $K_P = 1.7$ | $R_b = 15.6\ N$ | $R_c = 2\ N$ |
| Grade resistance | $R_g = 1471.5\ N$ | | | |

| | |
|---|---|
| Tractive effort and acceleration | $F = 3597.9\ N$ and $a = 5.6\ m/s^2$ |

## 4 Conclusion

The main contribution of this paper is to provide the key specifications following the standards ISO 7165:2009 and ISO 11601:2008, propose conceptual designs for the mobile manipulator for fire extinguisher testing, and design a tracked chassis that is suitable for the fire extinguisher testing application. In part II of this study, the process of designing, including manipulator modelling, control algorithms and simulation results, will continue to be discussed in detail.

## Acknowledgments

This research is funded by Vietnam National University Ho Chi Minh City (VNU-HCM) under grant number TX2022-20b-01. We acknowledge the support of time and facilities from National Key Laboratory of Digital Control and System Engineering (DCSELab), Ho Chi Minh City University of Technology (HCMUT), VNU-HCM for this study.